\documentclass[twoside,leqno,twocolumn]{article}

\usepackage[letterpaper]{geometry}

\usepackage{ltexpprt}
\usepackage{hyperref}
\usepackage{cite}
\usepackage{amsmath,amssymb,amsfonts}
\usepackage{graphicx}
\usepackage{booktabs}
\usepackage{subfigure}
\usepackage{algorithm}
\usepackage{xcolor}
\usepackage[utf8]{inputenc}
\usepackage[noend]{algpseudocode}

\begin{document}

\newcommand\relatedversion{}
\renewcommand\relatedversion{\thanks{The full version of the paper can be accessed at \protect\url{https://arxiv.org/abs/2207.01472}}} 

\title{\Large Deep Contrastive One-Class Time Series Anomaly Detection\relatedversion}
\author{Rui Wang \thanks{School of Computer Science and Engineering, Beihang University, Beijing, China.  \{ruiking, liucw, mxd\}@buaa.edu.cn, liuxd@act.buaa.edu.cn}
\and Chongwei Liu $^\dag$
\and Xudong Mou $^\dag$
\and Kai Gao\thanks{University of New South Wales, Sydney, Australia. kai.gao@unsw.edu.au}
\and Xiaohui Guo \thanks{Hangzhou Innovation Institute, Beihang University, Hangzhou, China. Xiaohui Guo is the corresponding author. guoxh@buaa.edu.cn}
\and Pin Liu \thanks{School of Information Engineering, China University of Geosciences in Beijing, Beijing, China. liupin@cugb.edu.cn} 
\and Tianyu Wo \thanks{College of Software, Beihang University, Beijing, China. woty@act.buaa.edu.cn}
\and Xudong Liu $^\dag$$^\S$}

\date{}

\maketitle


\fancyfoot[R]{\scriptsize{Copyright \textcopyright\ 2023 by SIAM\\
Unauthorized reproduction of this article is prohibited}}





\begin{abstract} \small\baselineskip=9pt 
The accumulation of time-series data and the absence of labels make time-series Anomaly Detection (AD) a self-supervised deep learning task. 
Single-normality-assumption-based methods, which reveal only a certain aspect of the whole normality, are incapable of tasks involved with a large number of anomalies. 
Specifically, Contrastive Learning (CL) methods distance negative pairs, many of which consist of both normal samples, thus reducing the AD performance.
Existing multi-normality-assumption-based methods are usually two-staged, firstly pre-training through certain tasks whose target may differ from AD, limiting their performance.
To overcome the shortcomings, a deep \textbf{C}ontrastive \textbf{O}ne-\textbf{C}lass \textbf{A}nomaly detection method of time series (COCA) is proposed by authors, following the normality assumptions of CL and one-class classification.
It treats the original and reconstructed representations as the positive pair of negative-sample-free CL, namely ``sequence contrast''.
Next, invariance terms and variance terms compose a contrastive one-class loss function in which the loss of the assumptions is optimized by invariance terms simultaneously and the ``hypersphere collapse'' is prevented by variance terms. 
In addition, extensive experiments on two real-world time-series datasets show the superior performance of the proposed method achieves state-of-the-art. 
\end{abstract}

\section{Introduction}
Within cyber-physical systems, sensor-equipped devices generate time-series data that contains massive status information, making it possible to detect unexpected errors and reduce maintenance costs in data-driven ways.
Anomaly Detection (AD) plays an increasingly important role in this context, which refers to detecting instances that are significantly dissimilar to the majority \cite{grubbs1969procedures}. 
Though the performance of deep learning methods is superior to shallow ones \cite{pang2021deep}, labeling the outlier from quantities of temporal data could be costly and tricky. 
So, AD is usually considered an unsupervised learning problem in which learning representation for discerning anomalies relies on some normality assumptions. 
For example, autoencoder-based \cite{malhotra2016lstm} methods assume normal samples are better restructured from the latent space than abnormal ones.
Similarly, one-class classification methods \cite{ruff2018deep} assume that the normal samples come from a single (abstract) class that could accurately describe the so-called “normality”.
However, these normality assumptions may be one-sided, some of which are just inspired by the pretext task of self-supervised representation learning.
Meanwhile, there are various time-series anomalies including point anomalies (global or local), subsequence anomalies, and anomaly time series \cite{blazquez2021review} (Fig. \ref{visual}), thus it is not sufficient to detect all based on one normality assumption alone. 

In particular, contrastive learning-based AD methods are emerging.
\cite{de2021contrastive} directly treats the InfoNCE loss of CPC \cite{oord2018representation} as the anomaly score for image AD, contrasting the context vector with the future representation vector.
NeuTraL AD \cite{qiu2021neural} devises a contrastive loss specific to a fixed set of learnable transformations and regards the training loss as the anomaly score, contrasting the transformed samples (views) with the original ones in the representation space.
The single-assumption-based CL AD methods above assume that more mutual information exists between normal comparison objects than anomalous ones. However, pairs transformed from different normal samples are treated as negative ones, pushing away many normal samples inside and not capturing shared information in the same class, similar to \cite{caron2020unsupervised}.
It goes against the very nature of AD, i.e., extracting features common to the vast majority of normal samples, thus leading to a decline in AD performance.

Indeed some scholars combine these normality assumptions into some compound ones to learn more expressive representations for downstream AD tasks.
For instance, Deep SVDD \cite{ruff2018deep} realizes a deep one-class classification framework for AD with deep features or representations learned by a pre-trained autoencoder.
\cite{sohn2020learning} presents the two-stage one-class classifier on contrastive representations and points out a subtle but important observation, i.e., the uniformity property of contrastive representation may hurt the one-class AD performance.
Even though, we argue that learning representation is distinct from capturing the normality and anomalies' underlying data regularities, formally they are two discrepant optimization objectives.
Therefore, with representation learning and outlier discriminating separated, the two-stage AD methods' performance is limited.
In addition, these AD methods are originally proposed in the computer vision domain, lacking temporal dependencies, thus generalizing them simply into time-series AD tasks is meaningless.

To address the above issues, we propose a one-stage negative-sample-free deep \textbf{C}ontrastive \textbf{O}ne-\textbf{C}lass \textbf{A}nomaly (COCA) detection model for time-series data. As shown in Fig. \ref{architecture},
first, the original training data is augmented, making it easier to isolate anomalies from normal samples.
Next, the augmented time series is encoded through a multi-layer temporal convolution neural network and then put into a Seq2Seq model in the latent space to learn the critical characteristics of time series, i.e., temporal dependencies. 
The key to CL is to pull contrasting objects (positive pairs) closer in the representation space, and researchers use a variety of positive pairs, such as context/future \cite{oord2018representation}, different augmentations \cite{chen2020simple}, and context/mask \cite{baevski2020wav2vec}.
Here, we regard the representation in the latent space and the representation reconstructed by the Seq2Seq model as positive pairs and name it ``sequence contrast". Note that it's different from an autoencoder, as the latter is a generative method, which performs the reconstruction of original data or so-called pixel-level generation\cite{chen2020simple}, carrying massive unnecessary details to downstream tasks.
Finally, the positive pairs are fed to a learnable nonlinear projection layer to obtain their projections respectively.

The model is trained via a contrastive one-class loss function with two terms: \textit{invariance} and \textit{variance}.
The invariance term is to maximize the cosine similarity between the one-class center, latent representations, and seq2seq outputs, instead of adjusting the hyper-parameters to balance the loss contribution of one-class and contrastive learning as in most multi-task learning.
The variance term is borrowed from \cite{bardes2022vicreg}, and the variance of the within-batch representations is maintained above a given threshold by a hinge loss to avoid ``hypersphere collapse'' without negative sample pairs, which also solves the difficulty of identifying negative pairs in AD. 
In practice, the invariance term is treated as the anomaly score for AD.
In conclusion, COCA combines the two normality assumptions that latent and reconstructed representations 1) have greater mutual information and 2) belong to a single class, without pre-training.
We summarize our contributions as follows:
\begin{itemize}
\item A novel normality assumption that combines CL and one-class classification for time-series AD.
\item A new time-series CL paradigm namely ``sequence contrast". By analyzing the problems solved with CL, we clarify that its essence is the representation, rather than the compared pairs or the negative examples.
\item A novel contrastive one-class loss function to optimize both contrastive learning and one-class classification, and avoid  ``hypersphere collapse" at the same time.
\item Extensive experiments performed on two datasets show that the proposed COCA leads to a new state-of-the-art in time-series AD.
\end{itemize}

\section{Related Work \label{related}}
This section contains a brief introduction of recent works in contrastive learning and deep anomaly detection. 

\textbf{Contrastive Learning.}
The recent renaissance of contrastive learning began with CPC \cite{oord2018representation} proposed InfoNCE, which pulls positive samples closer and distances negative samples, though relying on a large number of negative samples to learn a good representation.
\cite{wang2020understanding} summarizes two key properties of contrastive learning: 1) alignment: similar samples have similar representations (pull positive pair) and 2) uniformity: representations follow a uniform distribution on the hypersphere (push negative pair).
On the one hand, BYOL \cite{grill2020bootstrap}, SwAV \cite{caron2020unsupervised}, and SimSiam \cite{chen2021exploring} achieve uniformity in contrastive learning without using negative samples.
On the other hand, SimCLR \cite{chen2020simple} and TS-TCC \cite{eldele2021time} align augmented data representations to learn invariant representations for visual data and time series, respectively.
Also, TS-TCC uses a temporal contrasting module to address the temporal dependencies of time series.
Although all these contrastive learning approaches have successfully improved representation learning for visual data and time series, they could be inapplicable to time-series AD. 
For example, contradictions exist between the uniformity of contrastive learning and the class imbalance of anomaly detection.

\textbf{Deep Anomaly Detection.}
Recently, deep learning for anomaly detection has been regarded as a new research frontier of the AD field.
Deep anomaly detection methods can roughly be divided into two categories: deep learning for feature extraction and learning feature representations of normality \cite{pang2021deep}.
Deep learning for feature extraction is a two-staged learning method that uses deep methods to learn representations for downstream anomaly detection.
However, it does not directly address the anomaly detection task, so the representations learned in the pre-training may be detrimental to anomaly detection.
Learning feature representations of normality couples representations learning with anomaly scoring in some way, such as GANs-based \cite{schlegl2017unsupervised}, autoencoder-based \cite{malhotra2016lstm}, one-class classification-based \cite{ruff2018deep}, clustering-based \cite{zong2018deep}, saliency map-based \cite{ruff2018deep}, and contrastive learning-based \cite{qiu2021neural,de2021contrastive} methods.
The key to these methods lies in the assumption of normality/anomaly, and some assumptions of normality are inspired by the pretext task of self-supervised learning.
For instance, GANs-based methods assume normal samples are better generated from the latent space of the generative network than anomalies.
However, the normal sample assumption of these methods may explain only one aspect of overall normality, respectively.
Uniquely, COCA does not resort to pre-training and organically integrates the normality assumption of one-class classification and contrastive learning to detect anomalies for time-series data. 

\section{Methodology}
This section describes the proposed COCA in detail, including the structure, objective, and its relation to contrastive learning.

\subsection{Problem Definition.}
Given a set of time series $\mathcal{D}=\left\{{\bf X}_1,{\bf X}_2,\dots,{\bf X}_N\right\}$,
${\bf X}_i=\left\{x_1,x_2,\dots,x_T\right\}$ is a time series of length $T$,  where $x_{j}\in \mathop{\mathbb{R}}^{d}$ is a $d$-dimensional vector.
Since sliding windows are generally used to divide time series into length-$T$ sequences, $T$ has been called the sliding window length, as well.
$d=1$ means that the time series is univariate, and $d>1$ for multivariate.
In time-series AD, the anomaly score ${\bf S}_{i}$ of ${\bf X}_i$ is calculated by the AD model such that the higher ${\bf S}_{i}$ is, the more likely it is an anomalous time series.

\begin{figure}[h]
  \centering
  \includegraphics[width=\linewidth]{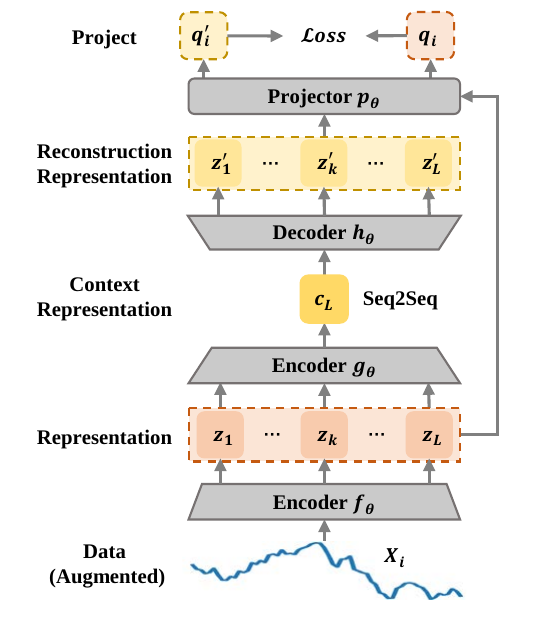}
  \caption{The overall architecture of the proposed COCA model.}
  \label{architecture}
\end{figure}

\subsection{Architecture.}
Fig. \ref{architecture} shows the architecture of the COCA model.
The time series ${\bf X}_i$ from an augmented training set of the raw dataset is passed to a multi-layer temporal convolution feature encoder $f_{\theta}:\mathcal{X}\mapsto\mathcal{Z}$ which takes as input time series ${\bf X}_i$ of length $T$ and outputs latent representations $z_{1},\dots, z_{L}$ for $L$ time-steps, potentially with a lower temporal resolution, i.e. $T>L$.
They are then fed to a Seq2Seq encoder $g_{\theta}:\mathcal{Z}\mapsto\mathcal{C}$ to summarize all $z_{\leq L}$ as context vectors $c_{L}$ and then a Seq2Seq decoder $h_{\theta}:\mathcal{C}\mapsto\mathcal{Z}$ produces reconstruction representations $z^{\prime}_{1},\dots, z^{\prime}_{L}$ for $L$ time-steps to learn temporal dependencies.
Furthermore, latent representations $z_{k}$ and seq2seq outputs $z^{\prime}_{k}$ are fed to a learnable nonlinear projector $p_{\theta}:\mathcal{Z}\mapsto\mathcal{Q}$ to output projections $q$ and $q^{\prime}$.
The output of the projector is used to calculate the loss (see next sub-section \ref{Objective}) to maximize the similarity between $q$ and $q^{\prime}$ concerning the one-class center $Ce \in \mathcal{Q}$ to combine the two normality assumptions: contrastive learning-based and one-class classification-based.

\textbf{Time-Series Augmentation.}
Data augmentation helps improve the performance of AD methods because it not only increases the volume of train data but also makes it easier to isolate anomalies \cite{sohn2020learning}.
In this paper, jittering (noise addition) and scaling (pattern-wise magnitude change) are applied to expand the training set.
Notably, the jittering and scaling hyper-parameters should be carefully chosen according to the nature of the time-series anomalies.

\textbf{Feature encoder.}
The encoder network has a 2-block temporal convolutional architecture, each block comprising a Conv1D layer, a BatchNormalization (BN) layer, a ReLU activation function, and a MaxPool1D layer, where the first block also contains a Dropout layer.
The time series input to the encoder should be normalized to zero mean and unit variance.

\textbf{Seq2Seq.}
The Seq2Seq consists of an encoder and a decoder. The encoder is a 3-layer Long Short-Term Memory (LSTM) and the decoder is a 3-layer LSTM followed by a  fully-connected (FC) layer.
In this paper, the hidden space representation length $L<20$, therefore LSTM can meet the needs of the context representation, while for long sequences, more recent advancements in Seq2Seq modeling such as self-attention networks or the Transformer model could help improve results further.

\textbf{Projector.}
The projector uses an MLP with one hidden layer applied BN and ReLU to map representations to the space where contrastive one-class loss is calculated. 

\subsection{The COCA Objective.\label{Objective}}
The COCA objective consists of \textit{invariance} and \textit{variance} terms.
The invariance term is to maximize the cosine similarity between the one-class center $Ce$, representations $q_{i}$, and seq2seq outputs $q_{i}^{\prime}$ in the projection space $\mathcal{Q}$, and the variance term avoids ``hypersphere collapse'' without negative sample pairs.

Before explaining the invariance term of the COCA objective, it is necessary to state the optimization objectives of one-class classification and contrastive learning without negative pairs.

\textbf{One-class classification.}
The optimization objective of Deep SVDD \cite{ruff2018deep}, a representative method for one-class classification, is defined as:
\begin{equation}
\mathcal{L}_{svdd} =\frac{1}{N}\sum_{i=1}^{N}\|\phi(x_{i},\Theta)-c\|^{2},
\label{Deep_SVDD_loss}
\end{equation}
where $c \in \mathcal{Z}$ is the one-class center, $\Theta$ is the set of parameters of a representation network $\phi$.
Deep SVDD obtains the sphere of the smallest volume by minimizing the $\mathcal{L}_{svdd}$ in the representation space $\mathcal{Z} \subset \mathop{\mathbb{R}}^{K}$.

\textbf{Negative-sample-free contrastive learning.}
BYOL \cite{grill2020bootstrap}, SimSiam \cite{chen2021exploring}, and Vicreg \cite{bardes2022vicreg} are representatives of contrastive learning without negative pairs.
The optimization objective of SimSiam is simplified as:
\begin{equation}
\mathcal{L}_{sim} =\frac{1}{N}\sum_{i=1}^{N}-\frac{z_{i}}{\Vert z_{i} \Vert_{2}} \cdot \frac{z_{i}^{\prime}}{\Vert z_{i}^{\prime} \Vert_{2}},
\label{cl_loss_2}
\end{equation}
where $z_{i}$ and $z_{i}^{\prime}$ are the representations of contrasting objects (positive pairs) in the latent space $\mathcal{Z}$.
Equation (\ref{cl_loss_2}) is essentially pulling the positive pair close using cosine similarity.
As for the ``hypersphere collapse" caused by no negative pairs, BYOL and SimSiam solve it by bootstrap and asymmetric networks, and Vicreg by variance.

\textbf{Invariance term of COCA objective.}
A crude way to integrate one-class classification and contrastive learning is treating it as multi-task learning with two adjustable hyper-parameters $\alpha$ and $\beta$ as follows:
\begin{equation}
\alpha \cdot \mathcal{L}_{svdd} + \beta \cdot \mathcal{L}_{sim}.
\label{multi_task}
\end{equation}
Therefore, the main intuition behind our model is that a positive correlation exists between one-class classification and contrastive learning, so their objectives can be achieved simultaneously by a loss function without hyper-parameters $\alpha$ and $\beta$.
Considering ${\rm sim}(u,v) = u^{T}v/\Vert u \Vert_{2}\Vert v \Vert_{2}$ denotes cosine similarity between $u$ and $v$, we define the invariance term $d$ between $\ell_{2}$-normalized $Q=\left\{q_1,q_2,\dots,q_N\right\}$ and $Q^{\prime}=\left\{q^{\prime}_1,q^{\prime}_2,\dots,q^{\prime}_N\right\}$ as:
\begin{equation}
d(Q,Q^{\prime}) = \frac{1}{N}\sum_{i=1}^{N}\left\{\left[1-{\rm sim}(q_{i},Ce)\right] + \left[1-{\rm sim}(q_{i}^{\prime},Ce)\right] \right\},
\label{invariance}
\end{equation}
where $Ce$ is the $\ell_{2}$-normalized one-class center defined by:
\begin{equation}
Ce(Q,Q^{\prime}) = \frac{1}{2N}\sum_{i=1}^{N}(q_{i} + q_{i}^{\prime}).
\label{center}
\end{equation}
Here, $Ce$, $q_{i}$, and $q_{i}^{\prime}$ are distributed on the unit hypersphere after normalization.
According to Equation (\ref{Deep_SVDD_loss}) $\mathcal{L}_{svdd}$, minimizing $d(Q,Q^{\prime})$ brings $q_{i}$ and $q_{i}^{\prime}$ closer to $Ce$, which achieves the one-class classification-based normality assumption.
Meanwhile, on the unit hypersphere, $d(Q,Q^{\prime})$ and $\mathcal{L}_{sim}$ are related as follows:
\begin{equation}
d(Q,Q^{\prime}) \geq 1+\mathcal{L}_{sim}(Q,Q^{\prime}),
\label{relation}
\end{equation}
which becomes tighter as $d(Q,Q^{\prime})$ decreases.
Also, observe that minimizing the $d(Q,Q^{\prime})$ shrinks an upper bound of contrastive errors $\mathcal{L}_{sim}(Q,Q^{\prime})$, and achieves the contrastive learning-based normality assumption. 
For more details see sub-section \ref{relation_cl}.

For the case where a little bit of training data is anomalous, which is very common in AD tasks, the \textit{soft-boundary invariance} of the COCA objective employing the hinge loss function is defined as:
\begin{equation}
d_{soft}(Q,Q^{\prime}) = L + \frac{1}{vN}\sum_{i=1}^{N}\max\left\{0,S_{i}-L\right\},
\label{soft_boundary_inv}
\end{equation}
where $L$ is the $(1-v)$-quantile of $S=\left\{S_1,S_2,\dots,S_N\right\}$, hyper-parameter $v\in(0,1]$ controls the trade-off between $L$ and violations of the boundary, i.e. the amount of time series allowed to be mapped outside the boundary.
$S_{i}$ is the \textit{anomaly score} of a time series ${\bf X}_i$, which is defined as:
\begin{equation}
S_{i}({\bf X}_i) = 2-{\rm sim}(q_{i},Ce)-{\rm sim}(q^{\prime}_{i},Ce),
\label{anomaly_score}
\end{equation}
where, $0<S_{i}({\bf X}_i)\leq 2$.

\textbf{Variance term of COCA objective.}
In AD, COCA removes negative pairs to avoid performance degradation caused by pushing away negative pairs that are both normal.
However, both negative-sample-free contrastive learning and Deep SVDD are likely to give an undesired trivial solution that all outputs “collapse” to a constant, i.e. ``hypersphere collapse''.
Inspired by \cite{bardes2022vicreg,chong2020simple}, COCA can then define the variance $v$ as a hinge function on the standard deviation of the projected vectors $q_{i}$:
\begin{equation}
v(Q) = \frac{1}{N}\sum_{i=1}^{N}\max\left\{0,\gamma - \sqrt{{\rm Var}(q_{i})+\varepsilon}\right\},
\label{variance}
\end{equation}
where $\gamma$ is a constant target value of the standard deviation, and $\varepsilon$ is a small scalar to prevent instabilities.
In our experiments, $\gamma$ is set to $1$, and $\varepsilon$ is set to $10^{-4}$.
On the other hand, according to the research in Deep SVDD, selecting an appropriate one-class center can alleviate the problem of hypersphere collapse.
In COCA, the one-class center $Ce$ is ensured to be non-zero in any dimension and only updated in the first few epochs, because experiments show that an unfixed $Ce$ would make the network easily converge to a trivial solution.

The overall loss function of COCA is a weighted average of the invariance and variance terms:
\begin{equation}
\mathcal{L} = \lambda d(Q,Q^{\prime}) + \frac{\mu}{2}(v(Q) + v(Q^{\prime})),
\label{coca_loss}
\end{equation}
where $\lambda$ and $\mu$ are hyper-parameters controlling the contribution of each term in the loss.
So similarly, the \textit{soft-boundary} loss function of COCA is defined as:
\begin{equation}
\mathcal{L}_{soft} = \lambda d_{soft}(Q,Q^{\prime}) + \frac{\mu}{2}(v(Q) + v(Q^{\prime})).
\label{coca_soft_loss}
\end{equation}
$\mathcal{L}$ applies to the training set without anomalies, while $\mathcal{L}_{soft}$ is for those containing a few anomalies.
Contrastive learning has two key properties: alignment and uniformity \cite{wang2020understanding} (detail in \ref{related}).
There is an inverse relationship between uniformity and hypersphere collapse, the better the uniformity the less likely the collapse will occur, and vice versa.
Nevertheless, uniformity somewhat contradicts the aim of one-class classification \cite{sohn2020learning}, because the latter is to bring representations closer to the center on the unit hypersphere, while some representations may be instead pulled far away by uniformity.
Therefore, in our experiments, $\lambda$ is fixed to $1$, and $\mu$ is determined by a grid search with the base condition $\mu < 1$.

\textbf{Anomaly Detection.}
In the test phase, an anomaly score $S_{i}$ will be generated for the time series ${\bf X}_i$.
Then, the following formula is applied to determine whether ${\bf X}_i$ can be classified as an anomaly:
\begin{equation}
\begin{aligned}  
x_{t}=\left\{
\begin{array}{lcl}
anomaly,& & S_{i} > \tau\\
 normal,& & S_{i} \leqslant \tau \quad ,
\end{array} \right.
\end{aligned}
\label{COCA_detection}
\end{equation}
where $\tau$ is a predefined threshold.
The overall algorithm is summarized in supplementary material A.2.

\begin{figure}[h]
  \centering
  \includegraphics[width=0.7\linewidth]{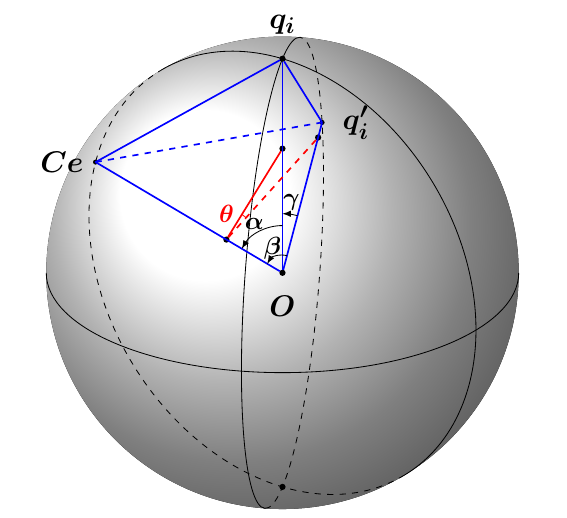}
  \caption{Invariance term schematic. $O$ is the center of the unit hypersphere, $Ce$ is the $\ell_{2}$-normalized one-class center, $q_{i}$ and $q_{i}^{\prime}$ are $\ell_{2}$-normalized projected vectors, $\theta$ is the dihedral angle between plane $CeOq_{i}$ and $CeOq_{i}^{\prime}$, $\alpha$ and $\beta$ are one-class errors, and $\gamma$ is the contrastive error.}
  \label{loss_visual}
\end{figure}

\subsection{Relation to Contrastive Learning. \label{relation_cl}}
COCA treats representations $q_{i}$ and reconstructed representations $q_{i}^{\prime}$ as positive pairs to learn shared information between different time steps of time series, discarding low-level information that is computationally expensive and unnecessary.
Along with CPC \cite{oord2018representation}, SimCLR \cite{chen2020simple}, and wav2vec \cite{baevski2020wav2vec}, though different in the types of positive pairs, COCA is essentially computing loss in the representation space.
Therefore, maximizing the cosine similarity of $q_{i}$ and $q_{i}^{\prime}$ in COCA is a type of negative-sample-free contrastive learning, and we name it ``sequence contrast''.
For time-series AD, COCA outperforms SimCLR-similar contrast methods that regard various augmentations as positive pairs (see sub-section \ref{main_result}).

Next, we will explain the mechanism of the invariance term to achieve the contrastive learning-based normality assumption.
As shown in Fig. \ref{loss_visual}, on the unit hypersphere, the angle $\alpha$/$\beta$/$\gamma$ is proportional to the Euclidean distance $l_{q_{i}Ce}$/$l_{q_{i}^{\prime}Ce}$/$l_{q_{i}q_{i}^{\prime}}$ between two points.
According to the triangle inequality, the relationship between the three Euclidean distances is $l_{q_{i}Ce} + l_{q_{i}^{\prime}Ce}\geq l_{q_{i}q_{i}^{\prime}}$. 
Therefore we are minimizing the cosine similarity between $q_{i}$, $q_{i}^{\prime}$, and $Ce$ in Equation (\ref{invariance}), which is an upper bound on the sequence contrastive learning errors between $q_{i}$ and $q_{i}^{\prime}$.
$\alpha$, $\beta$ and $\gamma$ are related as follows:
\begin{equation}
{\rm cos}\gamma = {\rm cos}\alpha{\rm cos}\beta + {\rm sin}\alpha{\rm sin}\beta{\rm cos}\theta,
\label{angle}
\end{equation}
where $\theta$ is the dihedral angle. 
According to Equation (\ref{angle}),
when $\alpha \to 0$ and $\beta \to 0$, ${\rm cos}\gamma \to 1$.
Therefore, Equation (\ref{relation}) becomes tighter as $d(Q,Q^{\prime})$ becomes smaller, which was also verified in our experiments.
For more formal proof details see supplementary material A.1.

\section{Experiments}
This section presents the experimental setup, baselines, COCA variants, main results, and hyper-parameter analysis.
The code is available at \url{https://github.com/ruiking04/COCA}.
\subsection{Experimental Setup.}
\textbf{Datasets.} 
Given the findings in \cite{wu2021current}, this paper abandons the flawed time-series AD datasets, such as Yahoo, Numenta, and NASA, and employs AIOps and UCR to evaluate the proposed model.
The datasets considered are as follows.
\begin{itemize}
    \item AIOps challenge (\textbf{AIOps}) \footnote{\url{https://github.com/NetManAIOps/KPI-Anomaly-Detection}}. This consists of well-maintained business cloud KPIs from some large Internet companies and contains 29 univariate time-series sub-datasets.
    \item UCR time series anomaly detection (\textbf{UCR})\footnote{\url{https://www.cs.ucr.edu/~eamonn/time_series_data_2018/}}\cite{UCRArchive2018}. This contains 250 univariate time-series sub-datasets from various fields.
\end{itemize}

Table \ref{datasets} summarizes these datasets.
The time series are partitioned into length-$T$ sequences by a sliding window with time-step $Ts\leq T$.
The two datasets both have a large number of samples and few anomalies, which is a challenge for some AD models.
This table shows the number of sequences in the training/validation/testing set, and the percentage of anomalous samples in the training/testing set.
The training sets of UCR don't contain anomalies, so the models are trained using Equation (\ref{coca_loss}), while \textit{soft-boundary} loss function Equation (\ref{coca_soft_loss}) is used for AIOps.

\begin{table}[!t]
\setlength{\abovecaptionskip}{0cm}
\setlength{\belowcaptionskip}{3pt}
\caption{Summary of time-series anomaly detection datasets}
\renewcommand{\arraystretch}{1}
\centering
\scalebox{0.86}{
\begin{tabular}{lcc}
\toprule
  & AIOps & UCR   \\ 
\midrule
 Number of sub-datasets  & 29  & 250\\
 Variables & 1 & 1 \\
 Domain & Cloud KPIs & Various\\
\midrule
 Length $T$  & 16  & 64 \\ 
 Time step $Ts$ & 2 & 4 \\
 Total samples  & 2961039  & 4830858 \\
\midrule
 Training/validation/testing & 40\%/10\%/50\% & 24\%/6\%/70\% \\
 Training/testing anomaly & 2.98\%/1.92\% & 0\%/0.71\%\\
\bottomrule
\end{tabular}}
\label{datasets}
\end{table}

\textbf{Evaluation Metrics.}
In most cases, time-series anomalies occur as continuous-time intervals rather than isolated points, leading to difficulty in quantifying the predicted anomaly label sequence. 
In recent years, many evaluation metrics for time-series AD have been proposed, such as NAB Score, Point-Adjusted (PA), Revised Point-Adjusted (RPA) metrics, etc., but these metrics may overestimate the performance of the AD algorithm \cite{kim2022towards}.
To achieve a rigorous evaluation of time-series AD, this paper uses two metrics: accuracy metric \cite{lu2022matrix} and affiliation metrics \cite{huet2022local}.
The $accuracy = n/250$ is a metric specifically for the UCR dataset, where $n$ is the number of correctly predicted sub-datasets.
Each sub-dataset in UCR contains only one anomaly segment, so as long as the predicted anomaly is within the correct region, this sub-dataset is considered correctly predicted.
Affiliation metrics calculate precision/recall/F1-score metrics based on the concept of “affiliation” between the ground truth and the prediction sets.
Note that affiliation metrics on the entire dataset are weighted averages of affiliation metrics for each sub-dataset:
\begin{equation}
{\rm F1_{entire}} = \sum_{i=1}^{M}\frac{k_{i}}{K}{\rm F1}_{i},
\nonumber
\end{equation}
where $M$ is the number of sub-datasets, $K$ is the total number of anomaly segments for the entire dataset, and $k_{i}$ is the number of anomaly segments of the $i$-th sub-dataset.

\subsection{Baselines and COCA Variants.}
The proposed approach is compared against the following unsupervised and self-supervised anomaly detection methods.

\textit{Traditional Anomaly Detection Baselines.} 
Three commonly used traditional anomaly detection baselines are adopted:
One-class SVM (OC-SVM)\cite{scholkopf1999support}, Isolation Forest (IF)\cite{liu2008isolation}, and Random Cut Forest (RCF)\cite{guha2016robust}. 

\textit{Deep Anomaly Detection Baselines.} Then, four deep anomaly detection methods: Deep one-class (Deep SVDD)\cite{ruff2018deep}, Spectral Residual CNN (SR-CNN)\cite{ren2019time}, Deep Autoencoding Gaussian Mixture Model (DAGMM)\cite{zong2018deep}, and LSTM Encoder-decoder (LSTM-ED)\cite{malhotra2016lstm}.

\textit{Contrastive Learning Anomaly Detection Baselines.} Finally, two contrastive learning baselines are set: Contrastive Predictive Coding Anomaly Detection (CPC-AD)\cite{de2021contrastive,oord2018representation} and Time Series Temporal and Contextual Contrasting Anomaly Detection (TS-TCC-AD) \cite{eldele2021time,sohn2020learning}. 

For Deep SVDD, we use Conv1D and LSTM to implement its autoencoder architecture to process time-series data.
Although DAGMM is initially designed for tabular data, in \cite{bhatnagar2021merlion} it is used for time-series data.
CPC is originally a method for sequential data, treating images as a sequence of pixels, so the network structure does not need to be changed significantly when processing time-series data.
For TS-TCC-AD based on \cite{sohn2020learning}, TS-TCC \cite{eldele2021time} is used to learn the representation of time series in the pre-training phase, and Deep SVDD is used for AD in the fine-tuning phase.

\textbf{COCA Variants.}
Moreover, we include the following five COCA variants as baselines to demonstrate the effectiveness of individual components in COCA.
\textit{NoAug} removes the time-series augmentations of COCA.
\textit{NoOC} removes the one-class classification of COCA to optimize the similarity of representations $q_{i}$ and reconstructed representations $q_{i}^{\prime}$. 
\textit{NoCL} removes the contrastive learning of COCA to optimize the similarity of representations and one-class center. 
The difference between the variant NoCL and Deep SVDD is that the former contains a learnable nonlinear projector $p_{\theta}$ network and no pre-training.
\textit{NoVar} removes the variance term of COCA to optimize the similarity of representations and one-class center. 
\textit{COCA-vi} treats different augmentations (jittering and scaling) as positive pairs for contrast learning, similar to SimCLR \cite{chen2020simple}.

\begin{table}[!h]
\setlength{\abovecaptionskip}{0cm}
\setlength{\belowcaptionskip}{3pt}
\caption{Average affiliation F1-score(\%) and accuracy (\%) with standard deviation for anomaly detection on time-series datasets. The best results are in bold.}
\renewcommand{\arraystretch}{1}
\centering
\setlength{\tabcolsep}{1mm}{
\scalebox{0.88}{
\begin{tabular}{c | c | c c}
     \toprule
	 \bfseries Datasets & \bfseries AIOps  &\multicolumn{2}{c}{\bfseries UCR}  \\ [0pt]
	 \midrule 
	  \bfseries Metric &  \bfseries Affiliation F1 & \bfseries Affiliation F1 & \bfseries Accuracy\\
	 \midrule
	 \bfseries OC-SVM & 25.36 & 60.26 & 8.80 \\
	 \bfseries IF & 33.24 & 59.40 & 37.60 \\
	 \bfseries RCF & 34.48$\pm$0.30 & 58.36$\pm$0.59& 38.67$\pm$0.68 \\
	 \midrule
	 \bfseries Deep SVDD & 38.23$\pm$0.65 & 37.19$\pm$1.35 & 7.60$\pm$1.73 \\
	 \bfseries SR-CNN & 31.54$\pm$1.03 & 51.72$\pm$0.83 & 30.40$\pm$0.91 \\
	 \bfseries DAGMM & 36.15$\pm$0.95 & 66.93$\pm$0.47 & 6.13$\pm$0.50 \\
	 \bfseries LSTM-ED & 34.12$\pm$0.54 & 66.87$\pm$1.07 & 51.02$\pm$2.05 \\
	 \midrule
	 \bfseries CPC-AD & 35.36$\pm$1.87 & 48.65$\pm$1.92 & 6.37$\pm$0.53 \\
 	 \bfseries TS-TCC-AD &  31.91$\pm$2.05 & 44.27$\pm$1.38 & 
 	 0.56$\pm$0.27 \\
	 \bfseries COCA & \textbf{66.78$\pm$2.91} & \textbf{79.16$\pm$1.27} &  \textbf{66.12$\pm$2.62} \\
	 \midrule
	  \bfseries NoAug &  65.74$\pm$4.61 & 57.24$\pm$1.35 & 26.64$\pm$2.35 \\
	  \bfseries NoOC & 51.49$\pm$5.96 & 62.33$\pm$2.05 & 33.96$\pm$2.61  \\
	 \bfseries NoCL & 63.80$\pm$3.29 & 77.80$\pm$1.82 & 63.84$\pm$3.65  \\
	 \bfseries NoVar & 65.90$\pm$2.45 & 78.82$\pm$1.60 & 65.16$\pm$2.81  \\
	 \bfseries COCA-vi & 65.86$\pm$3.07 & 75.48$\pm$1.20 & 60.36$\pm$2.25 \\

\bottomrule
\end{tabular}}}
\label{result}
\end{table}


\textbf{Implementation Details.}
The network structure of our proposed COCA consists of two parts: encoder and Seq2Seq.
The encoder comprises 2-block temporal convolutional modules that each are followed by batch normalization, ReLU activation, and $2\times 2$ max-pooling.
For the Seq2Seq, two identical three-layer LSTMs are employed with the same dropout rate at $0.45$ as 1D-CNNS.
As for optimizer, an Adam optimizer with a learning rate from $1e-4$ to $5e-4$, weight decay of $5e-4$, $\beta_{1} = 0.9$, and $\beta_{2} = 0.99$ is adopted.
On the AIOps dataset, after calculating the anomaly scores, COCA searches on anomaly sample rate $p$ from $0.01\%$ to $0.30\%$ with step $0.01\%$ to determine the optimal anomaly threshold $\tau$.
The UCR sub-datasets each have only one anomaly segment, so COCA directly takes the largest anomaly score as an anomaly.
In addition, for UCR we use the early stopping strategy, as the sub-datasets from different domains vary in epochs to convergence.
Each method is run 10 times to obtain the mean and standard deviation.
Lastly, all the models are built with PyTorch 1.7 and Merlion 1.1.1 \footnote{\url{https://github.com/salesforce/merlion}}\cite{bhatnagar2021merlion}, and trained on an NVIDIA Tesla V100 GPU.
See more details about augmentation and hyper-parameters in supplementary material B.2.

\subsection{Main Results. \label{main_result}}
We report affiliation F1-score and accuracy in Table \ref{result}.
From the vertical view of the table, some methods perform poorly on AIOps because there are anomalous samples in the training set, which leads to high false negative rates.
On the other hand, for the UCR dataset, methods such as OC-SVM, Deep SVDD, and TS-TCC-AD have higher F1-score but lower accuracy. That's because the accuracy metric is binary (anomaly found or not), and it indicates these methods' results are close to the correct range of ground-truth anomaly but do not fall within it.

From the horizontal view of the table, four conclusions can be drawn.
First, in shallow methods, RCF  with F1-score over $34\%$ performs well and even outperforms some deep methods, showing that RCF methods are good baselines in time-series AD.
Second, TS-TCC-AD and Deep SVDD each have a performance gap of over $7\%$ on the two datasets, indicating that regardless of pre-training methods, the pre-training process itself limits the performance.
It also further confirms that the pre-trained deep model limits the performance of two-staged AD methods.
Third, DAGMM and LSTM-ED perform better than other deep baselines, indicating the normality assumptions of clustering and reconstruction are more relevant to the nature of AD.
Last, the proposed COCA outperforms all baselines on both datasets, demonstrating the effectiveness and robustness of ensemble multiple normality assumptions.

Also, Table \ref{result} shows the effectiveness of each component in our proposed COCA model.
To be more specific, by analyzing the AD performance of the NoAug, augmentations improve the performance of AD on the two datasets, especially on UCR.
The results of COCA, NoOC, and NoCL show that the combination of multiple normality assumptions can improve the performance of AD effectively.
Meanwhile, the NoVar performs poorly compared to COCA, which makes it clear that the variance term of the COCA objective is important. 
The COCA-vi is $2\%$ averagely lower than the COCA on the two datasets because it treats different augmentations as positive pairs and ignores temporal dependencies.
Overall, the results of COCA are better than the five variants, indicating the effectiveness and necessity of each component in our model.

\begin{figure}[htb]
    \vspace{-0.3cm}  
    \centering
    \subfigure[AIOps]{
    \begin{minipage}[t]{0.5\linewidth}
    \centering
    \includegraphics[width=\linewidth]{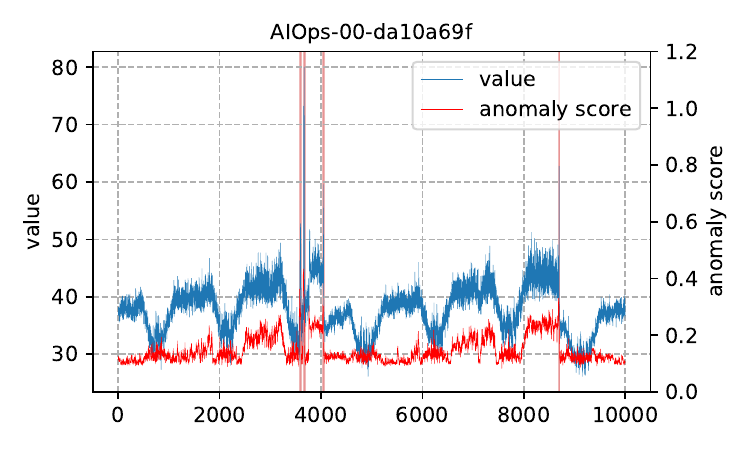}
    \label{visual-1}
    \end{minipage}
    }%
    \subfigure[UCR]{
    \begin{minipage}[t]{0.5\linewidth}
    \centering
    \includegraphics[width=\linewidth]{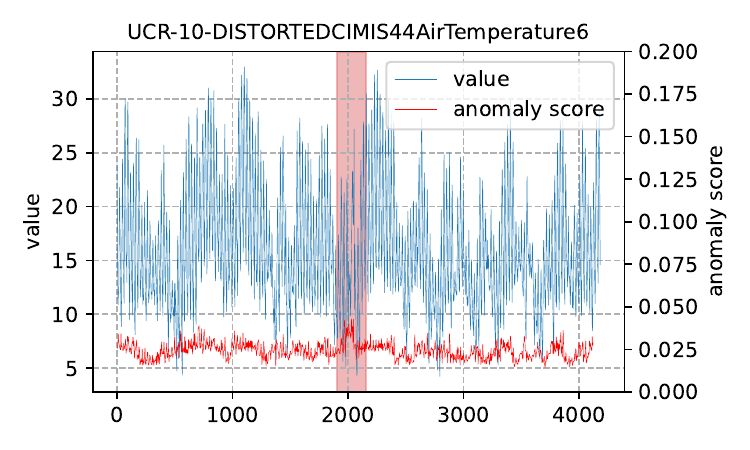}
    \label{visual-2}
    \end{minipage}
    }%
    \centering
    \caption{AD results of COCA on AIOps and UCR datasets. The blue curves are the original data value. The pink areas represent the ground-truth anomalies including point and subsequence anomalies. The red curves are the anomaly scores predicted by COCA.}
    \label{visual}
\end{figure}

\subsection{Visualization.}
To provide a more intuitive evaluation, visualizations of AD on AIOps and UCR are conducted, in Fig. \ref{visual}.
It can be seen that AIOps contains many point anomalies, which are suitable for some AD methods that are specialized in learning global features. 
In contrast, UCR contains both point and sequence anomalies.
COCA performs better on UCR compared to on AIOps, further illustrating that AD methods combining multiple normality assumptions can be applied to complex anomalous situations.

\subsection{Hyper-parameters Analysis.}
\begin{figure}[htb]
    \vspace{-0.3cm}  
    \centering
    \subfigure[$v$ (AIOps)]{
    \begin{minipage}[t]{0.5\linewidth}
    \centering
    \includegraphics[width=\linewidth]{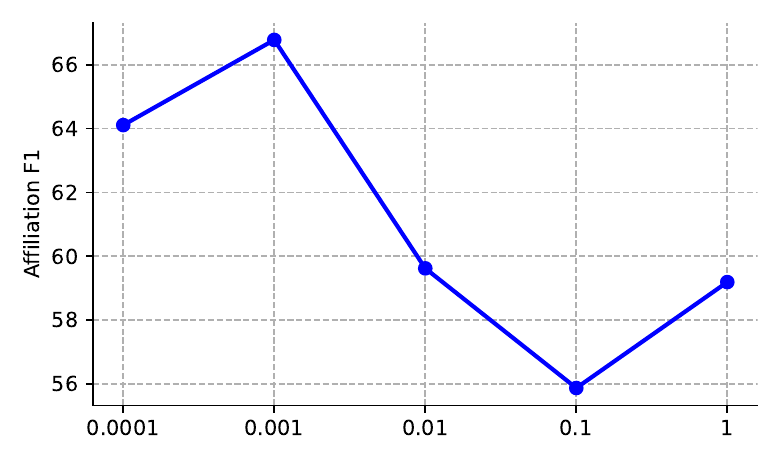}
    \label{para1}
    \end{minipage}
    }%
    \subfigure[$e$ (UCR)]{
    \begin{minipage}[t]{0.5\linewidth}
    \centering
    \includegraphics[width=\linewidth]{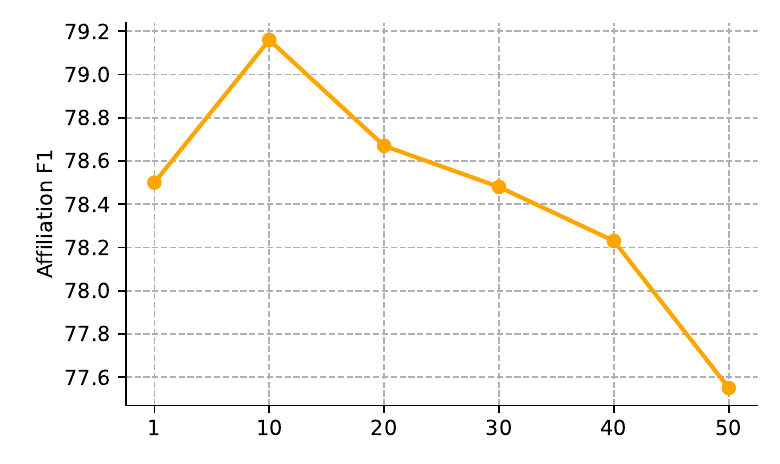}
    \label{para2}
    \end{minipage}
    }%
    \centering
    \caption{Two sensitivity analysis experiments on AIOps and UCR datasets. The left is the hyper-parameter $v\in(0,1]$ of \textit{soft-boundary invariance} and the right is training epoch $e$ before stopping updating the center $Ce$.}
    \label{hyper_parameters}
\end{figure}

In this section, sensitivity analysis is performed on the AIOps and UCR to study two main parameters: $v\in(0,1]$ in Equation (\ref{soft_boundary_inv}) and the epoch $e$ before stopping updating the center $Ce$.
Fig. \ref{para1} shows the effect of $v$ on the overall performance, where the y-axis is the affiliation F1-score metric.
For AIOps, we observe that $v=0.001$ is the best.
Apparently, appropriate anomaly proportions $v$ should be selected according to the anomaly proportion of datasets. 
Fig. \ref{para2} shows the results of varying epoch $e$ of stopping update center $Ce$ in a range between 1 and 50.
The model is shown to perform best on UCR when $e=10$, which suggests that the center $Ce$ should be frozen early since updating the center $Ce$ frequently increases the likelihood of hypersphere collapse.

\section{Conclusion}
We propose a novel deep framework called COCA for unsupervised time-series anomaly detection.
It combines the normality assumptions of contrastive learning and one-class classification, clarifies the essence of contrastive learning, and presents a new negative-sample-free type named ``sequence contrast''. Specially, we present a novel contrastive one-class loss function optimizing the loss of both assumptions simultaneously in one stage without tuning hyper-parameters as in most multi-task learning, as well as preventing ``hypersphere collapse''. Experiments on various datasets demonstrate that the performance of COCA achieves state-of-the-art.
We hope our work can help deepen the understanding of contrastive learning and offer more possibilities for fusion studies of various anomaly detection methods.

\section*{Acknowledgment}
This research is supported by the Zhejiang Province Key R\&D Program of China (2023C01070) and the Zhejiang Provincial Natural Science Foundation of China under Grant (LGG22F020043).
The authors acknowledge the providers of datasets, including AIOps and UCR \cite{UCRArchive2018}.

\bibliographystyle{siam}
\bibliography{reference}

\appendix
\section{Methodology Details}
This section provides the details and hyper-parameters for COCA time-series AD.

\subsection{Estimating the Invariance Term.}
As shown in Fig. \ref{loss_visual}, on the unit hypersphere, the formal proof is as follows:
\begin{equation}
\begin{aligned}
d(q_{i},q_{i}^{\prime})&=\left[1-{\rm sim}(q_{i},Ce)\right] + \left[1-{\rm sim}(q_{i}^{\prime},Ce)\right]\\
&\propto \alpha + \beta \\
&\propto l_{q_{i}Ce} + l_{q_{i}^{\prime}Ce} \\
&=\sqrt{\|q_{i}-Ce\|^{2}}+\sqrt{\|q_{i}^{\prime}-Ce\|^{2}} \\
&\geq \sqrt{\|q_{i}-q_{i}^{\prime}\|^{2}}\\
&\propto \gamma\\
&\propto 1-{\rm sim}(q_{i},q_{i}^{\prime})\\
&=1+\mathcal{L}_{sim}(Q,Q^{\prime}),
\end{aligned}
\label{prove}
\end{equation}
here, $l_{*}$ are the Euclidean distances. $\mathcal{L}_{sim}(Q,Q^{\prime})$ is the contrastive error expressing the agreement between positive pairs. 
\subsection{Detailed algorithms of COCA. \label{Algorithms}}
First, a pseudo-code for COCA in Pytorch style is provided in Algorithm \ref{pseudo}.

\renewcommand{\algorithmicrequire}{\textbf{Input:}}  
\renewcommand{\algorithmicensure}{\textbf{Output:}} 
\begin{algorithm}[htb]
  \caption{COCA‘s main training algorithm.} 
  \label{pseudo}  
  \begin{algorithmic}[0]
    \Require
      a set of augmented time series (jittering and scaling) $\left\{{\bf X}_{i}\right\}_{i=1}^{N}$,
      batch size $N$,
      structure of $f,g,h,p$,
      constant $nu,v, \gamma, \varepsilon, \lambda, \mu$.
    \Ensure
       Parameters of the network $f,g,h$, and $p$.
       \For{sampled batch $\left\{{\bf X}_{i}\right\}_{i=1}^{N}$}
            \ForAll{$i \in \left\{1,\dots,N\right\}$}
                   \\ \qquad \quad $\color{gray}\text{\# representations}$
                   \State ${\bf Z}_{i}=f({\bf X}_{i})$
                   \State $q_{i}=p({\bf Z}_{i})$
                   \\ \qquad \quad $\color{gray}\text{\# reconstruction representations}$
                   \State ${\bf Z}_{i}^{\prime}=h(g({\bf Z}_{i}))$
                   \State $q_{i}^{\prime}=p({\bf Z}_{i}^{\prime})$
                 \EndFor
           \State $Ce = \frac{1}{2N}\sum_{i=1}^{N}(q_{i} + q_{i}^{\prime})$
           \State \textbf{define} ${\rm sim}(u,v)$ \textbf{as} ${\rm sim}(u,v)=u^{T}v/\Vert u \Vert_{2}\Vert v \Vert_{2}$
           \ForAll{$i \in \left\{1,\dots,N\right\}$}
               \\ \qquad \quad $\color{gray}\text{\# anomaly score}$
               \State \textbf{define} $S_{i}({\bf X}_i)$ \textbf{as} $ 2-{\rm sim}(q_{i},Ce)-{\rm sim}(q^{\prime}_{i},Ce)$
            \EndFor
            \If{\emph{soft-boundary}}
                \State $L = {\rm quantile}(S({\bf X}), 1 - \eta)$
                \State $d(Q,Q^{\prime}) = L + \frac{1}{vN}\sum_{i=1}^{N}\max\left\{0,S_{i}-L\right\},$
            \Else
                \State $d(Q,Q^{\prime}) = \frac{1}{N}\sum_{i=1}^{N}S_{i}({\bf X}_i)$
            \EndIf
            \State $v(Q) = \frac{1}{N}\sum_{i=1}^{N}\max\left\{0,\gamma - \sqrt{{\rm Var}(q_{i})+\varepsilon}\right\}$
            \State $v(Q^{\prime}) = \frac{1}{N}\sum_{i=1}^{N}\max\left\{0,\gamma - \sqrt{{\rm Var}(q_{i}^{\prime})+\varepsilon}\right\}$
            \State $\mathcal{L} = \lambda d(Q,Q^{\prime}) + \frac{\mu}{2}(v(Q) + v(Q^{\prime}))$
            \\ \qquad update networks $f,g,h$, and $p$ to minimize $\mathcal{L}$ 
        \EndFor
        \Return network $f,g,h$, and $p$
  \end{algorithmic}
\end{algorithm}

\subsection{COCA Variants Loss Function.}
Moreover, we include the following five COCA variants as baselines to demonstrate the effectiveness of individual components in COCA.

\textit{NoAug.} The Variant NoAug removes the time-series augmentations of COCA.

\textit{NoOC.} The Variant NoOC removes the one-class classification of COCA to optimize the similarity of representations $q_{i}$ and reconstructed representations $q_{i}^{\prime}$. Its invariance term of the loss function is defined as:
\begin{equation}
\begin{aligned}
\frac{1}{N}\sum_{i=1}^{N}1-{\rm sim}(q_{i},q_{i}^{\prime}).
\end{aligned}
\label{NoOC}
\end{equation}

\textit{NoCL.} The Variant NoCL removes the contrastive learning of COCA to optimize the similarity of representations and one-class center. Its invariance term of the loss function is defined as:
\begin{equation}
\begin{aligned}
\frac{1}{N}\sum_{i=1}^{N}1-{\rm sim}(q_{i},Ce).
\end{aligned}
\label{NoCL}
\end{equation}
The difference between the variant NoCL and Deep SVDD is that the former contains a learnable nonlinear projector $p_{\theta}$ network and no pre-training.

\textit{NoVar.} The Variant NoVar removes the variance term of COCA to optimize the similarity of representations and one-class center. Its loss function is defined as:
\begin{equation}
\begin{aligned}
d(Q,Q^{\prime}).
\end{aligned}
\label{NoVar}
\end{equation}

\textit{COCA-vi.} The variant COCA-vi treats different augmentations (jittering and scaling) as positive pairs for contrast learning, similar to SimCLR \cite{chen2020simple}.
Its invariance term of the loss function is defined as:
\begin{equation}
\begin{aligned}
d(Z^{1},Z^{2}),
\end{aligned}
\label{COCA_vi}
\end{equation}
where $Z^{1}$ and $Z^{2}$ are the representations of the time-series data after jittering and scaling, respectively.

\section{Experiments.}
\subsection{Baseline Details. \label{Implementation1}}
For these deep baselines, Table \ref{baseline_tab} shows the normality assumptions, study domains, and whether two-staged.
\begin{table}[htb]
\setlength{\abovecaptionskip}{0cm}
\setlength{\belowcaptionskip}{2pt}
\caption{Summary of deep baselines.}
\renewcommand{\arraystretch}{1.3}
\centering
\scalebox{0.74}{
\begin{tabular}{lcccc}
\toprule
  & Assumption & Two-staged & Original domain   \\ 
\midrule
 Deep SVDD & Autoencoder\&One-class & $\surd$  & Image \\
 SR-CNN & Saliency map & $\times$ & Time series \\
 DAGMM & Clustering & $\times$  & Tabular data \\
 LSTM-ED & Autoencoder & $\times$  & Time series \\
 CPC-AD & Contrast & $\times$  & Time series \\
 TS-TCC-AD & Contrast\&One-class & $\surd$  & Time series \\
\bottomrule
\end{tabular}}

\label{baseline_tab}
\end{table}

\subsection{Hyper-parameters Details. \label{Implementation2}}
COCA is implemented in PyTorch, and some important parameter values used in the model are listed here, see Table \ref{parameter}.
In this table, \emph{repre\_channels} is the dimension of the final representations ${\bf Z}$, \emph{hidden\_size} is the dimension of the Seq2Seq in the model, and \emph{project\_channels} is the dimension of the projector.
\emph{window\_size} is the size of time window, the same as the length of time series $T$, and \emph{time\_step} is the step while sliding.
\emph{stop\_change\_center} is the training epoch $e$ before stopping updating the center $Ce$.
$\mu$ is the weight of the variance term of COCA objective.
\emph{lr} is the learning rate and \emph{nu} is the hyper-parameter $v\in(0,1]$ of \textit{soft-boundary invariance}. 
\emph{scale\_ratio} and \emph{jitter\_ratio} are the rate of scaling and jittering while applying data augmentation, respectively.


\begin{figure}[htbp]
  \setlength{\abovecaptionskip}{-0.15 cm}
\setlength{\belowcaptionskip}{0 pt}
  \vspace{0cm}  
  \centering
  \includegraphics[width=\linewidth]{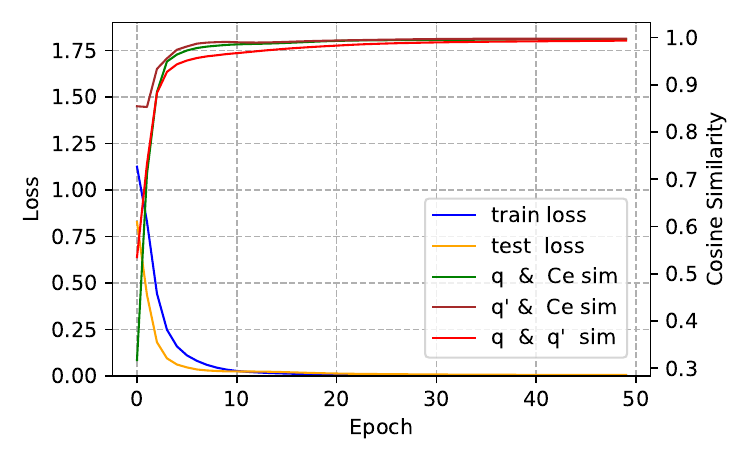}
    \caption{Loss and cosine similarity results for COCA and COCA-NoVar on UCR. Blue: train loss, Orange: test loss, Green: ${\rm sim}(q_{i},Ce)$, Brown: ${\rm sim}(q_{i}^{\prime},Ce)$, Red: ${\rm sim}(q_{i},q_{i}^{\prime})$.}
  \label{visual_var}
\end{figure}

\subsection{Relation to Contrastive Learning.}
To verify the validity of the invariance terms in the loss function of COCA, Fig. \ref{visual_var} illustrates loss and cosine similarity results for COCA on UCR.
As can be seen from Fig. \ref{visual_var}, the process of optimizing the loss function $\mathcal{L}$ makes ${\rm sim}(q_{i},Ce) \to 1$, ${\rm sim}(q_{i}^{\prime}, Ce) \to 1$ and ${\rm sim}(q_{i},q_{i}^{\prime}) \to 1$, which indicates that the loss we design not only makes $q_{i}$ and $q_{i}^{\prime}$ closer to $Ce$, but also minimizes the sequence comparison error ${\rm sim}(q_{i},q_{i}^{\prime})$.

\begin{table}[htb]
\setlength{\abovecaptionskip}{0cm}
\setlength{\belowcaptionskip}{2pt}
\caption{The values of hyper-parameters used in COCA}
\renewcommand{\arraystretch}{1.3}
\centering
\scalebox{0.88}{
\begin{tabular}{lcccccccc}
\toprule
  & AIOps & UCR  \\ 
\midrule
 repre\_channels  & 32  & 64  \\
 hidden\_size & 64 & 128 \\
 project\_channels & 16 & 32 \\
 window\_size  & 16  & 64 \\
 time\_step & 2 & 4  \\
 stop\_change\_center & 1 & 10 \\
 $\mu$ & 0.1 & 0.1 \\
 lr  & 0.0001 & 0.0003  \\
 nu & 0.001 & -  \\
 scale\_ratio & 1.1 & 0.8 \\
 jitter\_ratio & 0.1 & 0.2 \\
\bottomrule
\end{tabular}}
\label{parameter}
\end{table}

\end{document}